\documentclass{article} % For LaTeX2e
\usepackage{iclr2026_conference,times}

\usepackage{amsmath,amsfonts,bm}

\def\eqref#1{equation~\ref{#1}}
\def\1{\bm{1}}

\DeclareMathAlphabet{\mathsfit}{\encodingdefault}{\sfdefault}{m}{sl}
\SetMathAlphabet{\mathsfit}{bold}{\encodingdefault}{\sfdefault}{bx}{n}

\usepackage{hyperref}
\usepackage{url}
\usepackage{booktabs}
\usepackage{algorithm}
\usepackage{algorithmic}
\usepackage{graphicx}
\usepackage{multirow}%
\usepackage{amsmath,amssymb,amsfonts}%
\usepackage{amsthm}%
\usepackage{mathrsfs}%
\usepackage[title]{appendix}%
\usepackage{xcolor}%
\usepackage{textcomp}%
\usepackage{manyfoot}%
\usepackage{marvosym}
\usepackage{booktabs}%
\usepackage{listings}%
\usepackage{times}
\usepackage{epsfig}
\usepackage[table]{xcolor}
\usepackage{color}
\usepackage{ulem}
\usepackage{subcaption}

\title{GSPlane: Concise and Accurate Planar Reconstruction via Structured Representation}

\author{Ruitong Gan\textsuperscript{1,6}, Junran Peng\textsuperscript{2,6}, Yang Liu\textsuperscript{3,4,6}, Chuanchen Luo\textsuperscript{5,6}, Qing Li\textsuperscript{1}, \& Zhaoxiang Zhang\textsuperscript{3,4}\textsuperscript{\Letter} \\
\textsuperscript{1} The Hong Kong Polytechnic University \quad \textsuperscript{2} University of Science and Technology Beijing \\
\textsuperscript{3} NLPR, MAIS, Institute of Automation, Chinese Academy of Sciences\\
\textsuperscript{4} University of Chinese Academy of Sciences \quad \textsuperscript{5} Shandong University \quad \textsuperscript{6} Linketic \\
\texttt{ruitong.gan@connect.polyu.hk, jrpeng4ever@126.com} \\
\texttt{\{liuyang2022, zhaoxiang.zhang\}@ia.ac.cn}\\
\texttt{chuanchen.luo@sdu.edu.cn, qing-prof.li@polyu.edu.hk} \\
}

\iclrfinalcopy % Uncomment for camera-ready version, but NOT for submission.
\begin{document}

\maketitle

\begin{figure}[h]
  \centering
  \vspace{-0.3cm}
  % \fbox{\rule{0pt}{1.5in} \rule{\linewidth}{0pt}} % 调整这个值来减小虚拟盒子的垂直长度
  \includegraphics[width=1\linewidth]{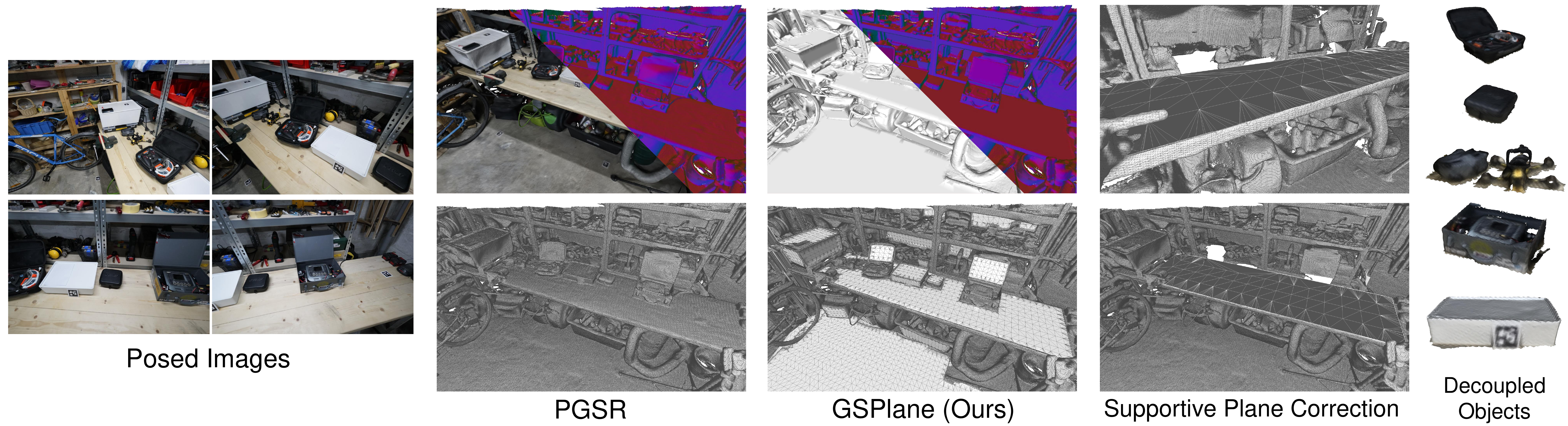} 
  %\vspace{-0.7cm} % 调整这个值来减小图与标题之间的垂直距离
  \captionof{figure}{We introduce GSPlane, which adopts 2D planar priors to constrain planar Gaussian distributions on corresponding planes. The structured representation for planes not only empowers the layout refinement of the mesh, resulting in topological correctness with notable less vertices, but also demonstrates the potential in decoupling objects and sealing contact regions on supportive planes.}
  \label{teaser}

\end{figure}

\begin{abstract}
Planes are fundamental primitives of 3D sences, especially in man-made environments such as indoor spaces and urban streets. Representing these planes in a structured and parameterized format facilitates scene editing and physical simulations in downstream applications. Recently, Gaussian Splatting (GS) has demonstrated remarkable effectiveness in the Novel View Synthesis task, with extensions showing great potential in accurate surface reconstruction. However, even state-of-the-art GS representations often struggle to reconstruct planar regions with sufficient smoothness and precision. To address this issue, we propose \textbf{GSPlane}, which recovers accurate geometry and produces clean and well-structured mesh connectivity for plane regions in the reconstructed scene. By leveraging off-the-shelf segmentation and normal prediction models, GSPlane extracts robust planar priors to establish structured representations for planar Gaussian coordinates, which help guide the training process by enforcing geometric consistency. To further enhance training robustness, a \textbf{Dynamic Gaussian Re-classifier} is introduced to adaptively reclassify planar Gaussians with persistently high gradients as non-planar, ensuring more reliable optimization. Furthermore, we utilize the optimized planar priors to refine the mesh layouts, significantly improving topological structure while reducing the number of vertices and faces. We also explore applications of the structured planar representation, which enable decoupling and flexible manipulation of objects on supportive planes. Extensive experiments demonstrate that, with no sacrifice in rendering quality, the introduction of planar priors significantly improves the geometric accuracy of the extracted meshes across various baselines.
\end{abstract}

\section{Introduction}

% 3D reconstruction from multi-view captured images is a widely-studied topic in both computer vision and graphics. High-precision reconstruction from images can help human understand and digitally depict related scenes, greatly reduce the burden of manual modeling, and have broad application prospects in areas such as robotics, virtual reality, video games, \textit{etc.} As a geometric concept, \textbf{planes} are one of the most frequently occurring structures in real scenarios, and also play a crucial role in simulating physical laws or objects interactions. Thus, reconstructing planes with proper geometry and topology is important in 3D scene reconstruction.
Planes are commonly witnessed in our daily environments, forming the foundation of many scenes: the streets and building facades outdoors, and the floors and ceilings indoors. When manually constructing digital assets, artists can easily leverage their priors knowledge to accurately model textures and geometric distributions in these areas. Establishing accurate planar structures not only enables concise meshes with considerably fewer vertices and faces, but also support downstream tasks such as physical simulation~\cite{qi2024shapellm}. In contrast, for example, if a reconstructed table is uneven or lacks geometric consistency, objects like cups would struggle to rest stably on its surface. Naturally, a question arises: in the context of creating digital twins through 3D reconstruction, can the introduction of planar priors help achieve more normal-consistent and accurate geometric reconstructions? Unfortunately, despite recent advancements in 3D reconstruction, there has been limited exploration of how to effectively leverage planar priors to address these challenges.

Recently, 3D Gaussian Splatting (3DGS)~\cite{kerbl20233d} introduced an explicit representation capable of achieving high-fidelity novel view synthesis in real time. Rather than relying on neural networks, 3DGS employs Gaussians characterized by parameters such as position, scale, rotation, color, and opacity. Its highly optimized rasterization pipeline enables fast rendering speed. Following 3DGS, several notable works~\cite{huang20242d, guedon2024sugar, zhang2024rade, yu2024gaussian, chen2024pgsr} focused on improving the Gaussian representation and depth regularization strategies to gain higher mesh quality in 3D surface reconstruction tasks, which has been extensively studied in the field of computer vision and graphics. Building on these advancements, methods such as GaussianRoom~\cite{xiang2024gaussianroom}, AGS-mesh~\cite{ren2024ags} further integrate prior information to enhance geometric accuracy. However, through our experiments, we observed that the prior knowledge in these methods is typically used as a supervisory signal to minimize regularization losses during training, and the Gaussian representations generated are not strictly constrained to lie on a single plane. Additionally, the meshing strategies adopted in these approaches tend to produce overly dense distributions of vertices and faces, especially for planar regions, leading to high-resolution demands that can be costly and less practical for downstream applications.

% To avoid inner depth ambiguity problem ~\cite{huang20242d} and improve geometric accuracy, many excellent works have been proposed. These methods either try to modify the volume of Gaussian spheres~\cite{huang20242d}, propose a regularization term to align the scene~\cite{guedon2024sugar}, replace projection-based rendering with ray-tracing-intersection with Gaussians~\cite{yu2024gaussian}, or estimate depth and surface normal for supervision~\cite{zhang2024rade}. Although the geometry of the mesh extracted from these methods looks similar to the original scene, there are errors in its topology between vertices, edges, and faces. The reconstructed mesh exhibits severe normal inconsistency in the plane region with the face distribution rough and rugged, which makes it difficult for further applications.

To address the aforementioned challenges, we propose \textbf{GSPlane}, a novel method that leverages planar priors from 2D images to generate meshes with consistent normal and coherent topology in planar regions. Our approach begins by estimating surface normal maps~\cite{hu2024metric3d} for each posed image and identifying potential planar regions using subpart mask proposals generated by SAM~\cite{kirillov2023segment}. These 2D planar priors are then projected into 3D space to cluster the initial 3D Gaussians into plane-specific groups. We introduce a structured representation for planar Gaussians by re-parameterizing their $xyz$ coordinates into a normalized weighted combination of three non-collinear basis points defining the plane. During training, both the basis points’ coordinates and the normalized weights for each planar Gaussian are optimized to refine the plane's orientation and position. To further improve accuracy, we incorporate a \textbf{Dynamic Gaussian Re-classifier} (DGR), which dynamically corrects false-positive planar Gaussians during training. The extracted mesh will be further refined by leveraging the optimized planar priors, enhancing the surface topology and layout in planar regions. Additionally, we explore \textbf{Supportive Plane Correction} (SPC), an applications of our structured planar representation, demonstrating its ability to improve mesh realism by preserving planar integrity and enabling flexible object manipulation across supportive planes.

% Before diving into the aforementioned problem, we should discuss some key knowledge about mesh and plane in advance. An ideal planar mesh with correct topology should have vertices aligned with the corresponding plane equation with sufficiently unified face normals. Existing mesh extractions fail to achieve this because they cannot identify the exact plane regions or unify face normals. Even methods that use surface normals from posed images for supervision cannot ensure complete normal consistency in planar regions during training. Therefore, it is necessary to distinguish planar and non-planar gaussian units while training the scene. An intuitive way is to adopt 2D planar priors derived from posed images. 

To thoroughly evaluate the effectiveness of 2D planar priors, we take both the indoor dataset ScanNetV2~\cite{dai2017scannet} and outdoor Tanks and Temples Dataset~\cite{knapitsch2017tanks} as benchmarks. Extensive experiments demonstrate that GSPlane achieves significantly better performance in planar regions, producing meshes with a unified layout and consistent normals—while maintaining rendering quality without any degradation. To summarize, the main contributions of the paper are:
\begin{itemize}
    \item We propose \textbf{GSPlane}, a powerful method that lifts 2D planar priors into 3D space and establishes a structured representation for planar Gaussians. Additionally, we incorporate optimized planar information during mesh layout refinement, ensuring topological correctness and consistency in the planar regions of the mesh.
    \item We present Supportive Plane Correction, an application of our structured planar representation that preserves planar integrity when decoupling objects from their supportive planes, enabling accurate planar geometry and facilitating flexible object manipulation. 
    \item Extensive experiments validate our SOTA surface reconstruction performance, showcasing promising benefits of 2D planar prior in 3D reconstruction.
\end{itemize}

\section{Related Works}
\subsection{Gaussian Splatting} 
% Recently, Gaussian Splatting (GS)~\cite{kerbl20233d} techniques have shown great potential in representing and reconstructing 3D scene information. With highly optimized rasterization pipeline, GS-based methods~\cite{ye2024geosplatting, wang2025freesplat, liang2024gus} achieve high-fidelity and real-time rendering of the reconstructed scene, as well as the mesh extraction techniques to obtain appropriate texture and geometry of the target scene. 2DGS~\cite{huang20242d} proposed to collapse the original 3D volume of gaussians into 2D to overcome the multi-view inconsistency of 3DGS, and provide better surface carving with depth distortion and normal consistency. Other methods~\cite{guedon2024sugar, yu2024gaussian, zhang2024rade} experimented different rasterization steps or splatting constraints to refine the scene representation from both the novel view synthesis and the extracted mesh. Some recent works~\cite{xiang2024gaussianroom, ren2024ags, turkulainen2024dn, wang2024gaussurf, dai2024high} further adopt surface normal and monocular depth information predicted from off-the-shelf models as additional supervisions in the training process, resulting in better surface reconstruction quality. However, the mesh extracted by these methods has poor smoothness and geometric accuracy in plane areas, resulting in a rugged surface and overly complex mesh structure.`z
Extracting accurate surfaces from unordered and discrete 3DGS is both a challenging and fascinating task. Numerous algorithms have been developed to extract high-quality surfaces while ensuring smoothness and managing outliers. The pioneering SuGaR~\cite{guedon2024sugar} approach pretrains 3DGS and integrates it with the extracted mesh for fine-tuning, utilizing the Poisson reconstruction algorithm for rapid mesh extraction. Techniques like 2DGS~\cite{huang20242d} and GaussianSurfels~\cite{dai2024high} reduce the original 3D Gaussian primitives to 2D to avoid ambiguous depth estimation. During GS training, the estimated normals derived from rendering and depth maps are aligned to ensure smooth surfaces. GOF~\cite{yu2024gaussian} focuses on unbounded scenes, using ray-tracing-based volume rendering to achieve a contiguous opacity distribution. RaDeGS~\cite{zhang2024rade} introduces a novel definition of ray intersection with Gaussian structures, deriving curved surfaces and depth distributions. Furthermore, recent works~\cite{xiang2024gaussianroom, ren2024ags, turkulainen2024dn, wang2024gaussurf, dai2024high, chen2024pgsr, zanjani2025planar, li2025mpgs, sun2025dreamcraft3d++} incorporate surface normal and monocular depth information predicted from off-the-shelf models as additional supervision in the training process, resulting in improved surface reconstruction quality and geometrical consistency. However, these mesh surfaces are still composed of overly dense distributions of vertices and faces, resulting in topological inaccuracies when compared to real-world structures. This excessive density not only leads to significantly larger file sizes but also poses challenges for subsequent editing and processing tasks.

\subsection{Traditional 3D Plane Reconstruction}
Traditional methods for 3D plane reconstruction often focus on identifying potential plane areas within a scene using RGB-D images~\cite{salas2014dense, silberman2012indoor, huang20173dlite} or sparse 3D point clouds~\cite{borrmann20113d, sommer2020planes}. By utilizing sets of points with 3D coordinates, either obtained from point clouds or derived from depth information, robust estimators such as PCA or RANSAC~\cite{fischler1981random} can be employed to fit geometric representations of planes. Other approaches~\cite{gallup2010piecewise, argiles2011dense} tackle the planar reconstruction problem through multi-view image segmentation, where each pixel is assigned to planar proposals represented in Markov Random Fields (MRF). In our research, we propose leveraging planar priors from 2D images to reconstruct target scenes. In earlier attempts, we proposed to directly post-process the reconstructed mesh~\cite{barda2023roar} via 2D planar priors, which led to significant errors in plane distribution. To address this, we introduced a structured planar representation that is optimized during training, allowing us to leverage learned plane equations to refine the reconstruction.

\subsection{Learnable 3D Plane Reconstruction} 

With the increasing availability of large-scale datasets containing both 2D images and 3D point clouds, learning-based methods have become the mainstream for extracting planar information from single images or videos. This capability facilitates the reconstruction of potential planes within a scene. Classical approaches, such as PlaneNet~\cite{liu2018planenet}, PlaneRecover~\cite{yang2018recovering}, and PlaneRCNN~\cite{liu2019planercnn}, segment possible plane distributions from a single image and optimize plane parameters using depth features to achieve a final reconstructed scene. PlanarRecon~\cite{xie2022planarrecon} is the first method to predict the planar representation of a scene from a sequence of images before reconstruction. Building on previous methods, Airplanes~\cite{watson2024airplanes} proposes estimating 3D-consistent plane embeddings and grouping them into scene instances. Uniplane~\cite{huang2024uniplane} uses sparse attention to query per-object embeddings for the scene. Alphatablets~\cite{he2024alphatablets} employs off-the-shelf surface normal and depth information to initialize small planes, which are further optimized to align with the scene's geometry and texture. While these methods show significant promise in reconstructing planar regions, they often produce less detailed and realistic geometric structures in non-planar areas. In contrast, out model well balance the performance in both planar and non-planar areas, achieving high quality for both rendering and surface reconstruction.

\section{Methods}\label{sec3}

\begin{figure*}[!ht]
\begin{center}
\includegraphics[width=0.98\textwidth]{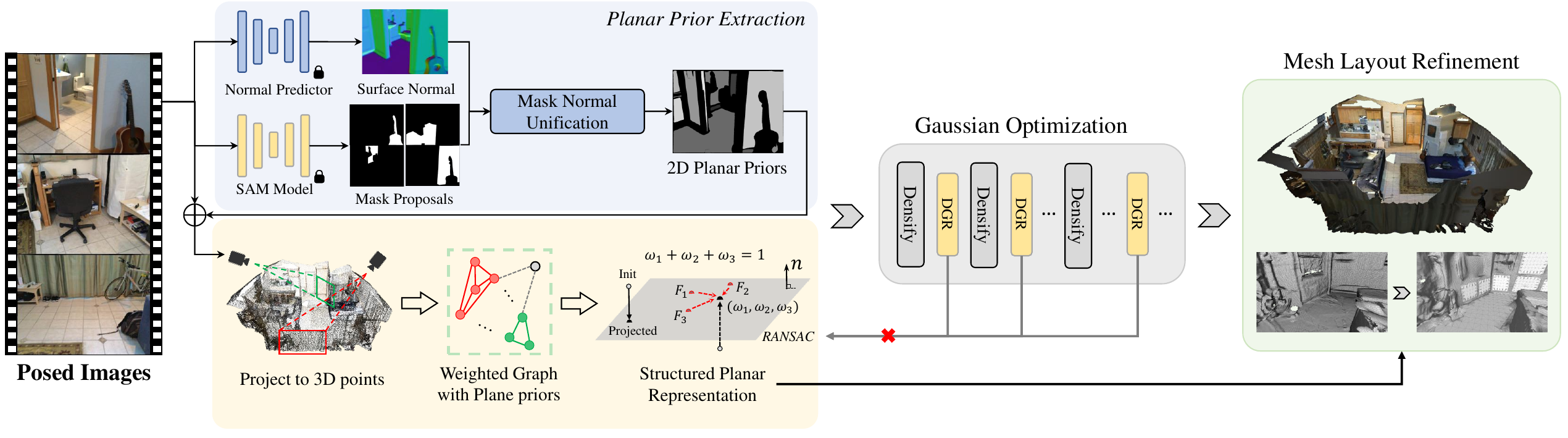} 
\end{center}
\vspace{-0.5cm}
\caption{Pipeline of \textbf{GSPlane}. Given a set of posed images as input, our method first extracts 2D planar priors from each view, align them with point cloud to obtain plane distributions in 3D space, and re-parameterize the related coordinates of Gaussians. During training, our Dynamic Gaussian Re-classifier continues to correct false-positive planar gaussian by reverting their representation back to $xyz$. The layout of the mesh extracted from training will also be refined with the optimized planar information from the structured representation.}

\label{vis:method}
\end{figure*}

Figure~\ref{vis:method} illustrates the overall pipeline of \textbf{GSPlane}. Starting with posed input images, \textbf{GSPlane} initially extracts planar prior information from each specific view, integrates them into the 3D point cloud, and establishes structured representations for 3D planar points before and during training (Sec.~\ref{sec3.1}). A \textbf{Dynamic Gaussian Re-classifier} (DGR) is then employed to refine the optimization process by identifying and correcting false-positive planar Gaussians (Sec.~\ref{sec3.2}). Finally, the extracted mesh is refined using the learned planar distributions to enhance surface topology and layout (Sec.~\ref{sec3.3}). Additionally, we propose \textbf{Supportive Plane Correction} (SPC), an application incubated from planar prior to improve realism by preserving planar integrity and enabling flexible object manipulation in reconstructed scenes (Sec.~\ref{sec3.4}).

\subsection{Structured Representation for Planes}\label{sec3.1}

Given a set of posed images $I_p = \{I_1, I_2, \dots, I_n\}$, potential planes are detected in each image using surface normal predictions.
For each image $I_i$, Metric3Dv2~\cite{hu2024metric3d} generates a surface normal map $N_i = (n_x, n_y, n_z)\in \mathbb R^{H\times W\times 3}$ and Segment-Anything-Model (SAM)~\cite{kirillov2023segment} produces subpart masks $M_i = \{M_{i,1}, M_{i,2}, \dots, M_{i,j}\}$ of the scene, where $i$ denotes the $i$-th image and $j$ for $j$-th mask.
For each mask region $M_{i,j}$,we compute the cosine similarity between the normals of individual pixels and the average normal of the region. If more than 70\% of the pixels in the region exceed the similarity threshold $\alpha$, these pixels are identified as a planar region. Overlapping planar regions are then merged into larger planar masks $P=\{P_1, P_2,...,P_n\}$ due to normal consistency.

Given an initial point cloud or COLMAP reconstruction of the scene, the coordinates of points in the point cloud are used to initialize the positions of the Gaussians. To incorporate planar priors into training, we establish planar relationships across different Gaussian units by projecting 2D planar masks from multiple views into 3D space.
we then construct an undirected graph $G(V, E)$, where each node $V_i$ corresponds to a point in the point cloud. 
An edge $E(V_i, V_j)$ is established between two nodes if the two corresponding points appear together on the same projected planar mask. The weight of the edge represents the frequency of these two points appearing on the same planar mask. Background points can be filtered out using depth information, and planar relationships are aggregated across all views on the graph $G$. 
The Leiden algorithm~\cite{traag2019louvain}, which is designed to detect communities in weighted graph, clusters nodes in $G$ into planar groups, and will be served as constraints for training Gaussians.
More details can be found in Appendix Sec.~\ref{Appendix:A}.

We assume that if a group of points $V_P$ in the point cloud is determined to lie on a plane, their corresponding Gaussian centers should also reside on the same plane. To impose this constraint, we introduce planar priors to re-parameterize Gaussian coordinates, replacing the direct optimization of $xyz$ positions with normalized weight parameters.
Specifically, for a planar cluster $V_P$, RANSAC~\cite{fischler1981random} is employed to estimate the plane, onto which all Gaussian centers are projected to obtain $V'_P$. From $V'_P$, three non-collinear points $F_1, F_2, F_3$ are randomly selected to serve as basis points defining the plane function. Each projected coordinate in $V'_P$ is then expressed as a normalized linear combination of the basis points:
\begin{equation}\label{eq1}
V'_P = \omega_1F_1 + \omega_2F_2 + \omega_3F_3, \quad \text{s.t.} \quad \omega_1 + \omega_2 + \omega_3 = 1.
\end{equation}
These weights $\omega_1, \omega_2, \omega_3$ are optimized during training to enforce planar constraints on the planar Gaussians.

\subsection{Dynamic Gaussain Re-classifier}\label{sec3.2}

Building upon the structured representation, planar Gaussians are optimized during training to adhere to planar constraints.
The coordinates of planar Gaussians, whether initialized directly from the point cloud or derived through densification, are represented using basis points and normalized weights (Eq.~\ref{eq1}). While the coordinates of the basis points are optimized as well, they are assigned a lower learning rate to allow for adjustments in plane orientation and position.

The accuracy of the planar Gaussian relations and the effectiveness of the planar priors are closely tied to the performance of SAM and Metric3Dv2. However, in cases where a Gaussian is misclassified as planar (\textit{i.e.}, a false-positive planar Gaussian), it cannot be correctly optimized according to the planar coordinate formulation in Eq.~\ref{eq1}. To address this issue, we propose the \textbf{Dynamic Gaussian Re-classifier} (DGR) to identify and reclassify such false-positive planar Gaussians. During the DGR phases, gradients for both planar and non-planar Gaussians are collected and averaged for evaluation. The top 5\% of planar Gaussians, based on their average gradients, are then compared to the average gradient magnitude of the top 20\% of non-planar Gaussians. If the gradient magnitude of a planar Gaussian exceeds the average gradient magnitude of the top 20\% non-planar Gaussians, the coordinates of that planar Gaussian are re-formulated back into the $xyz$ coordinate format. DGR operates iteratively between Gaussian densification and after the final densification step. The implementation details are provided in Sec.~\ref{Appendix:B} in the Appendix.

\subsection{Mesh Layout Refinement}\label{sec3.3}

Traditional mesh generation methods applied after Gaussian Splatting often produce overly dense meshes with redundant vertices and faces, which not only reduce geometric accuracy but also compromise storage efficiency.To address this, we introduce a mesh layout refinement procedure that leverages planar priors to optimize mesh structure in planar regions. This refinement improves normal consistency, topological coherence, and reduces vertex density, facilitating object decoupling from supportive planes like floors or tables.

Starting from an initial mesh $O$ (e.g., generated via TSDF~\cite{curless1996volumetric}, Marching Tetrahedra~\cite{shen2021deep}, etc.), we first identify clusters of mesh vertices that correspond to known planar regions. These planar relationships are precomputed from the sparse point cloud $Pcd$ as sets $V_P^i$, where $i$ indexes the $i$-th detected plane. We transfer these planar relationships from the point cloud to the mesh by assigning mesh vertices to planes using a spatial proximity criterion based on the voxel size $\delta$.
Specifically, for a given plane $A$, a mesh vertex $v_x \in O$ is considered to belong to plane $A$ if:
\begin{equation}
    \{v_x ~| ~\exists{v_y} \in V_P^A, ~|v_y - v_x| < 1.5\delta ~~\wedge ~~ \forall{\bar{v}_y} \notin V_P^A, ~|\bar{v}_y-v_x|>0.5\delta \}.
\end{equation}
This ensures that each mesh vertex is matched to a unique planar region with sufficient spatial confidence.

Once planar vertex clusters are identified, we refine each planar region individually. We begin by removing all mesh faces formed by three vertices lying on the same plane, retaining only the associated vertices. Each planar vertex cluster is then classified into two categories. \textbf{Boundary vertices}, which are connected via mesh edges to vertices outside the planar cluster and thus form the perimeter of the planar region. \textbf{Interior vertices}, which are fully enclosed within the planar region and have no direct connections to non-cluster vertices. 
Both boundary and interior vertices are projected onto their corresponding planes, defined by the optimized basis points. To regularize the interior structure, we replace interior vertices with a set of uniformly distributed 2D grid points on the plane. These grid points serve as candidates for reconstructing the triangulated surface of the planar region.
However, we observe that planar regions in meshes often have irregular shapes, which can cause misalignment between the grid layout and the actual geometry. To mitigate this, we compute the minimum enclosing rectangle (MER) of the projected vertices. The MER provides a consistent 2D bounding frame aligned with the local plane axes, enabling uniform placement of grid points along the $x$- and $y$-directions. 
Considering the actual region of the plane in mesh, grid points falling outside the projected planar region are discarded. The remaining grid points, together with the projected boundary vertices, form a 2D point setthat is triangulated using Delaunay triangulation~\cite{lee1980two}.This produces a set of triangular faces that seamlessly connect the planar interior to its boundary. Finally, the 2D grid coordinates and their associated faces are mapped back into 3D space using the plane basis, and the resulting vertices and faces are integrated into the original mesh. This results in a refined planar region with consistent normals, reduced redundancy, and improved geometric structure. The complete mesh refinement algorithm is detailed in Alg.~\ref{alg:mesh_refinement} in the Appendix.

\subsection{Supportive Plane Correction}\label{sec3.4}

Conventional mesh reconstruction methods often merge individual objects and structural elements into a single, overly connected surface. This results in unrealistic geometry, particularly in regions where objects are in contact. For instance, when attempting to digitally separate an object - such as removing a cup from a table - the reconstructed mesh may exhibit gaps or voids in the contact area, failing to preserve the original physical continuity of the supporting surface.
To address this challenge, we propose leveraging planar priors to refine mesh representations within designated planar regions. This approach, referred to as \textbf{Supportive Plane Correction} (SPC), is an optional refinement step in our method designed to handle planar surfaces that serve as object-supporting structures, such as tables, shelves, or floors. To address this issue, we introduce an optional refinement step termed \textbf{Supportive Plane Correction} (SPC), which leverages planar priors to improve mesh representations of object-supporting surfaces, such as tables, shelves, or floors. Unlike general planar regions, supportive planes typically exhibit structural incompleteness — characterized by multiple internal voids (e.g., holes within the plane) or missing boundary regions (e.g., incomplete edges). 
SPC builds upon the mesh layout refinement process described in Sec.~\ref{sec3.3}, with key modifications tailored to preserve the integrity of supportive planes. Specifically, during grid point sampling, points that fall outside the initially projected planar region are \textit{retained} rather than discarded. In contrast, boundary vertices that define voids or holes are \textit{excluded} from the Delaunay triangulation step.This ensures that the resulting triangulated surface spans the full extent of the plane while avoiding reintroducing known discontinuities.
Beyond structural refinement, SPC enables flexible and physically plausible object manipulation. By isolating and sealing the contact regions between objects and their supporting surfaces, individual objects can be repositioned or removed without affecting the geometry of the underlying plane. This capability enhances both the visual realism and editability of the reconstructed scene by preserving planar surface continuity while enabling object-level interaction.

% \begin{table}[!h]
% \centering
% \tabcolsep=10pt
% % \resizebox{\columnwidth}{!}{%
% \begin{tabular}{l|ccccc}
% \toprule
% Setting     & Acc\downarrow    & Comp\downarrow  & Prec\uparrow  & Recall\uparrow & F-score\uparrow \\ \midrule
% \rowcolor{gray!20}
% 2DGS~\cite{huang20242d}        & 0.0661 & 0.0782 & 0.6035 & 0.5676  & 0.5834   \\
% 2DGS + normal   & 0.0645 & 0.0764 & 0.6396 & 0.5972  & 0.6177   \\
% 2DGS + Ours-train       & \textbf{0.0630} & \textbf{0.0733} & \textbf{0.6501} & \textbf{0.6197}  & \textbf{0.6330}   \\ \midrule
% \rowcolor{gray!20}
% RaDe-GS~\cite{zhang2024rade}      & 0.1008 & 0.1041 & 0.4805 & 0.5069  & 0.4914   \\
% RaDe-GS + normal & \textbf{0.0947} & 0.1024 & \textbf{0.5179} & 0.5388  & 0.5281   \\
% RaDe-GS + Ours-train     & 0.0960 & \textbf{0.1016} & 0.5069 & \textbf{0.5576}  & \textbf{0.5283}   \\ \bottomrule
% \end{tabular}%
% % }
% \vspace{-0.2cm}
% \caption{Ablation on normal utilization. We calculate L2-loss between estimated normals from these two methods and normals from Metric3Dv2.}
% \vspace{-0.5cm}
% \label{tab:3}
% \end{table}

% 这里是原本的大型可视化figure，占位

% \begin{figure*}[!h]
% \begin{center}
% %\fbox{\rule{0pt}{2in} \rule{0.9\linewidth}{0pt}}
% \includegraphics[width=0.95\textwidth]{vis1_small.pdf} 
% \end{center}
% \vspace{-0.4cm}
% \caption{Visualization of the reconstruction performance on scenes from ScanNetV2~\cite{dai2017scannet}. We provide comparisons with four baseline methods in Gaussian Splatting. `Ours' means that we apply both the SPR and PME on the baselines.}
% \label{vis1}
% \end{figure*}

\section{Experiments}
\subsection{Experimental Settings}
\noindent \textbf{Dataset.} We conduct extensive experiments on both the indoor dataset ScanNetV2~\cite{dai2017scannet} and outdoor dataset Tanks and Temples Dataset~\cite{knapitsch2017tanks}. Both datasets provides ground-truth mesh for evaluation. We evaluate scenes in terms of geometric accuracy, plane-wise geometric accuracy, and rendering quality compared with previous methods. 

\noindent \textbf{Metrics.} To evaluate the \textbf{scene-wise} geometric reconstruction performance, we follow the protocol of PlanarRecon~\cite{xie2022planarrecon} and report metrics including \textit{Accuracy}, \textit{Completion}, \textit{Precision}, \textit{Recall}, and \textit{F-score}. Additionally, we adopt the approach from Airplanes~\cite{watson2024airplanes} to report \textbf{planar-wise} metrics such as \textit{fidelity}, \textit{completion}, and \textit{L1 chamfer}. These metrics are evaluated on the $k=20$ and $k=30$ largest planes sampled from ground truth mesh using PlaneRCNN~\cite{liu2019planercnn}. Note that planar-wise metrics can only be assessed on meshes produced through our Planar-Guided Mesh Extraction, as baseline methods do not incorporate planar information in the extracted mesh. Please refer to airplanes~\cite{watson2024airplanes} for more details. To comprehensively evaluate performance, we also provide metrics about rendering quality, including PSNR, SSIM, and LPIPS, as done in 3DGS~\cite{kerbl20233d}.

\noindent \textbf{Implementation Details} We implement our GSPlane method on five representative GS-based methods, including 3DGS~\cite{kerbl20233d}, 2DGS~\cite{huang20242d}, GOF~\cite{yu2024gaussian}, RaDe-GS~\cite{zhang2024rade}, and PGSR~\cite{chen2024pgsr}. The initial mesh is extracted with the proposed process from the baseline, with the voxel size as 0.005. 
Note that the Marching Tetrahedral used in GOF closes all boundaries, including the ceilings of indoor scenes and empty plane regions, which violate the actual mesh distribution. Thus, when introducing our strategy to GoF, we abort this technique and turn to TSDF fusion for mesh extraction, so as to avoid mesh in actually empty areas. During the experiment, we set the threshold of cosine similarity $\alpha$ to 0.98.

\subsection{Overall Performance}
The indoor quantitative results of the overall metrics are presented in Tab.~\ref{tab:1}. Specifically, \textit{Ours-train} denotes applying structured representation of planes and Dynamic Gaussain Re-classifier in the training stage, while \textit{Ours-full} further incorporates mesh layout refinement in the post-training stage. Note that the \textbf{Supportive Plane Correction} (SPC) step is excluded from the performance evaluation. For a fair comparison, we also report results from GaussianRoom~\cite{xiang2024gaussianroom} and AlphaTablets~\cite{he2024alphatablets}, which leverage normal maps, depth, and edge information as priors for reconstruction. Compared with the methods that adopt off-the-shelf predictions for direct supervision, our GSPlane demonstrates the effectiveness of incorporating planar priors. The results highlight that the structured plane representation consistently improves both geometric and rendering quality across baselines, while the proposed mesh layout refinement enables more accurate and complete surface estimation. \textit{Ours-train} achieves a slight reduction in vertex count compared to baseline methods because it produces tighter and more compact planar distribution of Gaussians, while \textit{Ours-full} significantly reduces the number of vertices in the final mesh. Notably, the structured Gaussian planar representation also contributes to enhanced rendering quality, see Sec.~\ref{Appendix:C} in Appendix for rendering visualizations.

\begin{table*}[!t]
    \centering
    \tabcolsep=15pt
    \resizebox{\textwidth}{!}{
    \begin{tabular}{l|ccccc|ccc|c}
    \toprule
    \multicolumn{1}{c|}{}                                        & \multicolumn{5}{c|}{Geometry}                                                       & \multicolumn{3}{c|}{NVS}                           & Mesh            \\ \hline
    Method                                                       & Acc$\downarrow$  & Comp$\downarrow$ & Prec$\uparrow$   & Recall$\uparrow$ & F-score$\uparrow$ & SSIM$\uparrow$  & PSNR$\uparrow$    & LPIPS$\downarrow$ & Vertices        \\ \midrule
    \rowcolor{gray!20} GaussianRoom~\cite{xiang2024gaussianroom} & 0.084          & 0.062          & 0.602          & 0.621          & 0.611           & 0.779          & 23.89           & 0.36            & 3.01M          \\
    \rowcolor{gray!20} Alphatablets~\cite{he2024alphatablets}    & 0.094          & 0.219          & 0.501          & 0.446          & 0.459           & -              & -               & -               & 139.4K          \\ \midrule
    3DGS~\cite{kerbl20233d}                                      & \uline{0.083}    & 0.099          & 0.453          & 0.429          & 0.436           & 0.849          & 23.494          & 0.321           & 2.24M          \\
    3DGS + Ours-train                                            & 0.088          & \uline{0.097}    & \uline{0.459}    & \uline{0.438}    & \uline{0.446}     & \textbf{0.853} & \textbf{23.718} & \textbf{0.320}  & 2.00M          \\
    3DGS + Ours-full                                             & \textbf{0.077} & \textbf{0.080} & \textbf{0.471} & \textbf{0.656} & \textbf{0.548}  & -              & -               & -               & \textbf{1.23M} \\ \midrule
    2DGS~\cite{huang20242d}                                      & 0.066          & 0.078          & 0.603          & 0.568          & 0.583           & 0.845          & 22.673          & 0.346           & 1.73M          \\
    2DGS + Ours-train                                            & \uline{0.063}    & \uline{0.073}    & \uline{0.650}    & \uline{0.620}    & \uline{0.633}     & \textbf{0.847} & \textbf{23.263} & \textbf{0.337}  & 1.60M          \\
    2DGS + Ours-full                                             & \textbf{0.058} & \textbf{0.062} & \textbf{0.664} & \textbf{0.716} & \textbf{0.689}  & -              & -               & -               & \textbf{946.1K} \\ \midrule
    GOF (Tetra.) ~\cite{yu2024gaussian}                                & 0.120          & 0.111          & 0.413          & 0.484          & 0.444           & 0.810          & 21.444          & \textbf{0.357}  & 41.7M           \\
    GOF (TSDF) + Ours-train                                          & \uline{0.100}    & \uline{0.091}    & \uline{0.477}    & \uline{0.598}    & \uline{0.528}     & \textbf{0.828} & \textbf{22.460} & 0.359           & 1.89M          \\
    GOF (TSDF) + Ours-full                                           & \textbf{0.086} & \textbf{0.080} & \textbf{0.482} & \textbf{0.686} & \textbf{0.566}  & -              & -               & -               & \textbf{1.02M} \\ \midrule
    RaDe-GS~\cite{zhang2024rade}                                 & 0.101          & 0.104          & 0.480          & 0.507          & 0.491           & 0.829          & 22.334          & \textbf{0.348}  & 1.49M          \\
    RaDe-GS + Ours-train                                         & \uline{0.096}    & \uline{0.101}    & \uline{0.507}    & \uline{0.558}    & \uline{0.528}     & \textbf{0.832} & \textbf{22.394} & 0.351           & 1.45M          \\
    RaDe-GS + Ours-full                                          & \textbf{0.082} & \textbf{0.086} & \textbf{0.520} & \textbf{0.674} & \textbf{0.587}  & -              & -               & -               & \textbf{794.3K} \\ \midrule
    PGSR~\cite{chen2024pgsr}                                 & 0.079          & 0.085          & 0.581          & 0.571          & 0.573           & 0.847          & 25.350          & 0.274  & 5.3M          \\
    PGSR + Ours-train                                         & \uline{0.065}    & \uline{0.063}    & \uline{0.633}    & \uline{0.640}    & \uline{0.634}     & \textbf{0.852} & \textbf{25.494} & \textbf{0.261}           & 5.2M          \\
    PGSR + Ours-full                                          & \textbf{0.062} & \textbf{0.059} & \textbf{0.636} & \textbf{0.658} & \textbf{0.646}  & -              & -               & -               & \textbf{2.9M} \\ \bottomrule
    \end{tabular}
    }
    \vspace{-0.2cm}
    \caption{Quantitative evaluations including both the overall geometric scores and novel view synthesis (NVS) metrics on ScanNetV2~\cite{dai2017scannet} scenes. `Ours-train' denotes applying structured representation for planes and DGR. `Ours-full' denotes additionally applying mesh layout refinement after training.}
    
    \label{tab:1}
\end{table*}

\begin{table*}[t]
\centering
\renewcommand{\arraystretch}{1.15}
\resizebox{\textwidth}{!}{%
\begin{tabular}{l|cc|cc|cc|cc|cc}
\toprule
Metric                     & 3DGS   & 3DGS + Ours & 2DGS   & 2DGS + Ours & GOF    & GOF + Ours & RaDe-GS & RaDe-GS + Ours & PGSR & PGSR + Ours \\ \midrule
F-score$\uparrow$                         & 0.09   & \textbf{0.17}        & 0.32   & \textbf{0.34}        & 0.46   & \textbf{0.47}       & 0.40   & \textbf{0.42} & \textbf{0.52} & 0.52        \\ 
Planar Vertices               & 317.5K & \textbf{4.53K}       & 609.3K & \textbf{6.94K}       & 3.04M  & \textbf{41.26K}     & 503.1K  & \textbf{6.89K} & 2.39M & \textbf{29.27K}      \\ 
Overall Mesh Vertices        & 1.86M  & \textbf{1.55M}       & 3.75M  & \textbf{3.03M}       & 57.82M & \textbf{53.58M}     & 2.39M & \textbf{1.76M} & 14.69M  & \textbf{12.04M}      \\ \bottomrule
\end{tabular}%
}

\caption{Quantitative evaluations on Tanks and Temples Dataset~\cite{knapitsch2017tanks}.}
\vspace{-0.4cm}
\label{tab:tnt}
\end{table*}

\begin{figure*}[t]
    \centering
    \vspace{-0.8cm}
    \includegraphics[width=0.85\textwidth]{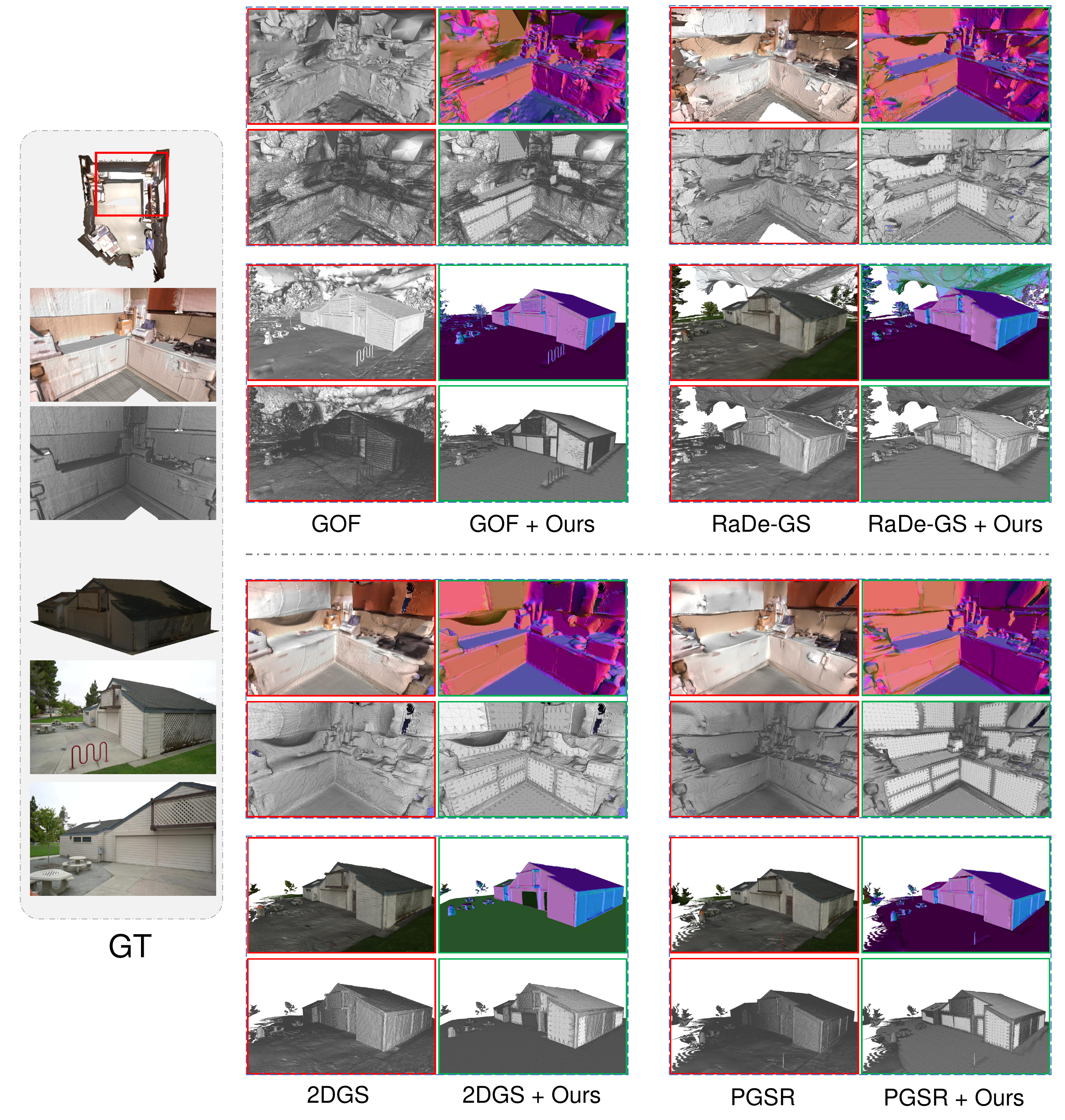}
    \vspace{-0.3cm}
    \caption{Visualizations of the mesh performance on both indoor and outdoor scenes. We provide comparisons on four baseline methods. It can be seen from the refined normal map and wireframes that our method can reduce the number of vertices by large margin, while maintaining consistent normal and topology across different planes. More examples can be found in Appendix.}
    \label{fig:overall}
\end{figure*}

The outdoor quantitative results are displayed in Tab.~\ref{tab:tnt}, where we report the F-score as the reconstruction metric, along with the number of planar and total vertices for comparison. \textit{Ours} in Tab.~\ref{tab:tnt} corresponds to the \textit{Ours-full} configuration  in Tab.~\ref{tab:1}. As seen in the table, our method improves reconstruction performance in outdoor scenes while significantly reducing the number of vertices in the mesh. However, the geometric improvements are less pronounced compared to indoor scenes, primarily because the TNT dataset contains fewer planar regions in some scenarios compared to ScanNetV2. Nevertheless, our method still achieves substantial reductions in mesh vertex count, demonstrating its efficiency in outdoor settings. Visualizations for both indoor and outdoor scenes can be found in Fig.~\ref{fig:overall} and Fig.~\ref{appendix:vis} in the Appendix.

\subsection{Planar-wise Geometry}
The planar metrics, including Fidelity, Accuracy, and L1-Chamfer Distance, are presented in Tab.\ref{tab:planar}. Our proposed planar-guided mesh extraction demonstrates significant potential for improving the reconstruction of planar regions across various Gaussian Splatting baselines. More visualizations on processing planar priors and mesh quality comparison can be found in Appendix Sec.~\ref{Appendix:D}.

\begin{table}[!t]
\centering
\tabcolsep=25pt
\resizebox{0.8\columnwidth}{!}{%
\begin{tabular}{l|ccc}
\toprule
Method          & Fidelity$\downarrow$ & Acc$\downarrow$  & CD$\downarrow$   \\ \midrule
\rowcolor{gray!20}PlanarRecon~\cite{xie2022planarrecon}     & 18.86     & 16.21 & 17.53 \\
\rowcolor{gray!20}AirPlanes~\cite{watson2024airplanes}       & 8.76      & 7.98  & 8.37  \\
\rowcolor{gray!20}PlanarSplatting~\cite{tan2025planarsplatting} & 6.64      & 11.76 & 9.2   \\ \midrule
3DGS + Ours-full         & 6.21 / 6.75      & 7.95 / 8.15  & 7.08 / 7.35  \\
2DGS + Ours-full          & \textbf{5.49} / 5.82      & 7.32 / 7.71 & 6.41 / 6.77 \\
GOF + Ours-full           & 8.25 / 8.74     & 9.50 / 9.93 & 8.88 / 9.34 \\
RaDe-GS + Ours-full       & 7.57 / 7.83     & \textbf{6.34} / \textbf{6.60} & 6.96 / 7.22 \\
PGSR + Ours-full       & 5.24 / \textbf{5.39}      & 6.58 / 6.65  & \textbf{5.91} / \textbf{6.02}  \\ \bottomrule
\end{tabular}%
}
\vspace{-0.2cm}
\caption{Planar-wise metrics evaluated on $k=20 / k=30$ largest plane regions from gt mesh in ScanNetV2, following Airplanes~\cite{watson2024airplanes}. The results from methods displayed in grey are evaluated with $k=20$ from the papers.}

\label{tab:planar}
\end{table}

\subsection{Ablation Study}
\begin{figure*}[!t]
\begin{center}
\includegraphics[width=0.95\textwidth]{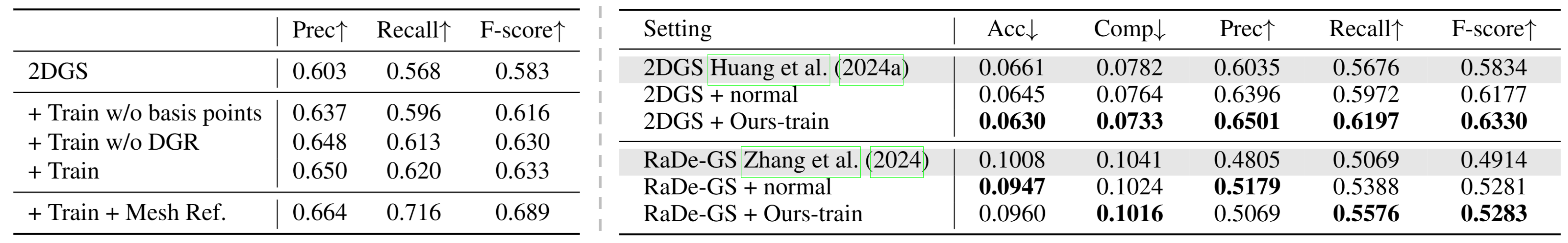} 
\end{center}

\caption{Ablation study results on GSPlane. The left table shows the effectiveness of different modules in GSPlane, and the right table compares our structured representation with off-the-shelf normal map supervision for mesh geometry reconstruction.}
\label{Fig:ablation}
\vspace{-0.5cm}
\end{figure*}

 We conduct an ablation study to evaluate the effectiveness of different modules in GSPlane, including the optimization of basis points, the \textbf{Dynamic Gaussian Re-classifier} (DGR), and the post-refinement of the mesh layout. The results are presented in the left table of Fig.~\ref{Fig:ablation}. Compared to the baseline performance of 2DGS, our GSPlane significantly enhances the quality of the generated mesh. Additionally, we perform experiments on 2DGS and RaDe-GS, both of which estimate normal maps during the rasterization process. Our goal is to analyze the differences between our proposed structured representation and directly using off-the-shelf normal maps to supervise the estimated normals. As shown in the right table of Fig.~\ref{Fig:ablation}, adopting our structured representation leads to better geometric performance in the reconstructed mesh. For ablation studies on hyperparameters, please refer Tab.~\ref{tab:alpha} and Tab.~\ref{tab:sigma} in Appendix Sec.~\ref{Appendix:A}.

%  这是nvs效果，占位

% \begin{figure}[h]
% \begin{center}
% %\fbox{\rule{0pt}{2in} \rule{0.9\linewidth}{0pt}}
% \includegraphics[width=1\linewidth]{nvs.pdf} 
% \end{center}
% \vspace{-0.3cm}
% \caption{Visualization of NVS results. Our Structured Planar Representation can reduce floaters by constraining planar Gaussians, resulting in visual improvements in some baseline methods.}
% \vspace{-0.5cm}
% \label{nvs}
% \end{figure}

%  这是zoom in视角展示，占位
% \begin{figure}[h]
% \begin{center}
% %\fbox{\rule{0pt}{2in} \rule{0.9\linewidth}{0pt}}
% \includegraphics[width=1\linewidth]{vis2.pdf} 
% \end{center}
% \vspace{-0.5cm}
% \caption{Visualization of the Planar-Guided Mesh Extraction results. (a) Baseline mesh. (b) Mesh extracted from non-planar gaussians. (c) Mesh after Planar-Guided Mesh Extraction (PME). (d) Wireframe of plane in baseline mesh. (e) Wireframe of PME mesh. Our method obtain structured plane representation with correct geometry and unified normal vectors.}
% \vspace{-0.2cm}
% \label{vis2}
% \end{figure}

\subsection{Application on Supportive Plane}

\begin{figure}[htbp]
    \centering
    \vspace{-0.3cm}
    \includegraphics[width=0.95\textwidth]{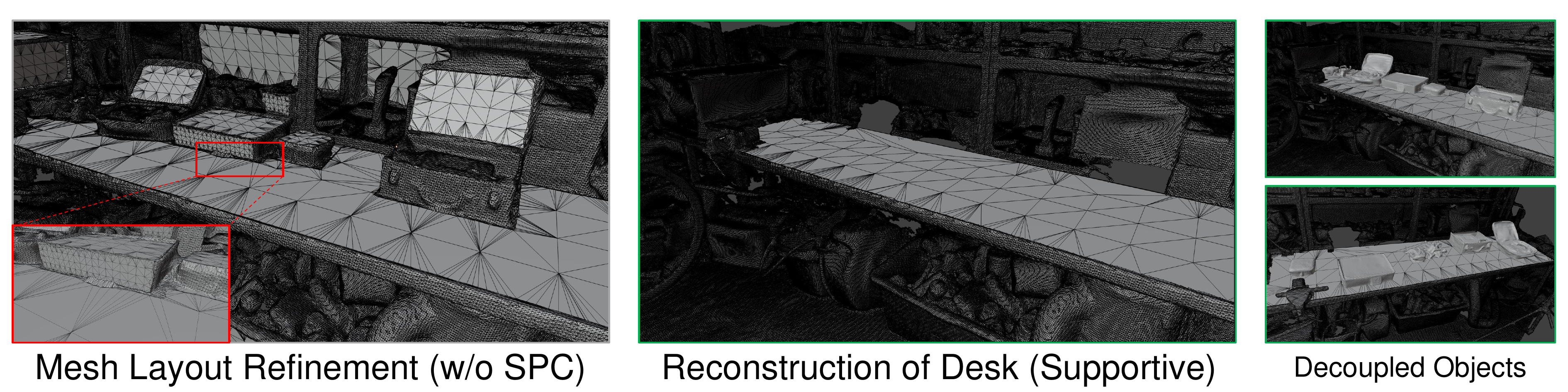}
    \vspace{-0.2cm}
    \caption{Visualizations of Supportive Plane Correction. When running SPC, the object boundaries are excluded from plane reconstruction, leading to an intact plane with complete shape like in reality. The objects are decoupled from the supportive plane surface, and can be further moved or manipulated freely.}
    \label{fig:supportive}
    \vspace{-0.5cm}
\end{figure}

To validate the effectiveness of \textbf{Supportive Plane Correction} (SPC), we conducted experiments demonstrating its ability to accurately reconstruct supportive planes and decouple objects resting on them. As shown in the left of Fig.~\ref{fig:supportive}, the default result of mesh layout refinement can provide unified grid points on plane, but the boundaries of the placed object are connected with the grid points to maintain wholeness of the structure. By fully utilizing the optimized planar priors, it is possible to infer the real shape and structure of the supportive plane - desk, and objects placed on the desk can also be removed from the desk. This ensures that the reconstructed supportive plane remains continuous and free of artifacts, even in the presence of complex void geometries.
The hole of the objects at the contact area can also be sealed using the supportive plane function, and are further free to manipulate across the supportive plane or within the scene.

\section{Conclusion}

In this paper, we highlight the potential of incorporating plane prior knowledge into Gaussian Splatting for improved reconstruction of planar regions. By leveraging segmentation and surface normal estimation, GSPlane generates structured planar representations, improving the geometric accuracy and topological consistency of meshes while reducing the density of vertices and faces. Additional discussion on supportive plane demonstrates that our structured planar representation enables realistic plane completion and decouples objects from planes, allowing further object manipulation.
Our experiments demonstrate that leveraging this prior significantly enhances the geometric accuracy and topological consistency of extracted meshes, reducing the complexity of the mesh structure.

\bibliography{iclr2026_conference}
\bibliographystyle{iclr2026_conference}
\clearpage
\appendix

\section{Additional Illustration on Structured Representation for Planes}\label{Appendix:A}

This section provide additional illustrations on how the structured representation for planes are obtained from 2D images. After obtaining the normal maps and subpart mask proposals from off-the-shelf models, we first multiply the normal map $N_i$ with each mask proposal $M_{i,j}$, where $i$ denotes the $i$-th image and $j$-th mask, to isolate the normal distribution $N_{mask}$ within each instance region. To determine if a region is planar, we take cosine similarity to measure the distance between each pixel normal to the average normal within the instance. Empirically, if more than 70\% of the pixels have a similarity larger than a certain threshold $\alpha$, we regard these pixels as a single plane. The largest connected region of these valid pixels is then selected as a plane proposal. In case multiple planes are mistakenly segmented into a single mask and do not meet the previous condition, we apply K-means clustering to the normals in this region with pixel number bigger than $\sigma$. We then evaluate each cluster using the 70\% criterion to identify all potential planes. If none of the clusters meet the criterion, the mask proposal is considered non-planar. In our experience, setting the target number of clusters to 2 yields good results. By following these steps, we can identify all the plane proposals $M^i_{plane}$ in image $I_i$.

To address the potential intersections among the obtained plane proposals $\mathcal{M}^i_{plane}$, we implement a series of steps to resolve conflicts in these overlapping areas. We first define an empty list $\mathcal{M}^i_{merge}$ to store the exclusive planar masks after the process. We iteratively select each element $M^{i,k}_{plane}$ in $\mathcal{M}^i_{plane}$, and compute normal vector cosine similarity with all other proposals $M^{i,l}_{plane}$. If any proposals matches through aforementioned 70\% criteria, they are merged together with $M^{i,k}_{plane}$ and pop out from $\mathcal{M}^i_{plane}$. The final $M^{i,k}_{plane}$ will be stored in $\mathcal{M}^i_{merge}$. After completing all the planar proposals in $\mathcal{M}^i_{plane}$, we achieve a collection of mutually exclusive planar masks $\mathcal{M}^i_{merge}$. By assigning each element with an index, we are able to obtain the final planar mask $P_i$. The overall algorithm is detailed in Alg.~\ref{alg1}.

\begin{algorithm}
\caption{2D Planar Perception}
\begin{algorithmic}[1]
\REQUIRE $\text{normal map } N_i, \text{mask proposals }\{M_{i,j}\}$

\FOR{each $M_{i,j}$}
    \STATE $N_{\text{mask}} \leftarrow N_i \odot M_{i,j}$,
    \STATE $d \leftarrow cos\_sim(N_{\text{mask}}, \overline{N_{\text{mask}}})$
    \IF{$\text{ratio}(d > \alpha) < 0.3$} 
        \STATE $M^{i,j}_{plane} \leftarrow M_{i,j}[d > \alpha]$
    \ELSIF{$\text{Area}(N_{mask}) > \sigma$}
        \STATE $N_{cluster1}, N_{cluster2} \leftarrow \text{K-means}(N_{mask})$
        \STATE $\text{Repeat Step 2-5 on } N_{cluster1}, N_{cluster2}$
    \ENDIF
\ENDFOR
\STATE $\mathcal{M}^i_{plane} \leftarrow [M^{i,j}_{plane}]$
\STATE $\mathcal{M}^{i}_{merge} \leftarrow \text{empty list}$
\WHILE{$\mathcal{M}^i_{plane} \text{ not empty}$}
    \STATE $M^{i,k}_{\text{plane}} \leftarrow \mathcal{M}^i_{plane}[0]$
    \FOR{each $l \neq k$}
        \STATE $d' \leftarrow cos\_sim(\overline{M^{i,k}_{\text{plane}}}, \overline{M^{i,l}_{plane}})$
        \IF{$d' > \alpha$}
            \STATE $M_{\cap} \leftarrow M^{i,k}_{\text{plane}} \cap M^{i,l}_{plane}$
            \STATE $M^{i,k}_{\text{plane}} \leftarrow M^{i,k}_{\text{plane}} + M^{i,l}_{plane} - M_{\cap}$
            \STATE $\mathcal{M}^i_{plane}.\text{pop}(M^{i,l}_{\text{plane}})$
        \ENDIF
    \ENDFOR
    \STATE $\mathcal{M}^{i}_{merge}.push(M^{i,k}_{\text{plane}})$
\ENDWHILE

\STATE $P_i \leftarrow  \text{assign instance ID with } \mathcal{M}^{i}_{merge}$
\RETURN {$P_i$}
\end{algorithmic}
\label{alg1}
\end{algorithm}

\begin{figure}[h]
\vspace{-0.3cm}
\begin{center}
\includegraphics[width=0.65\linewidth]{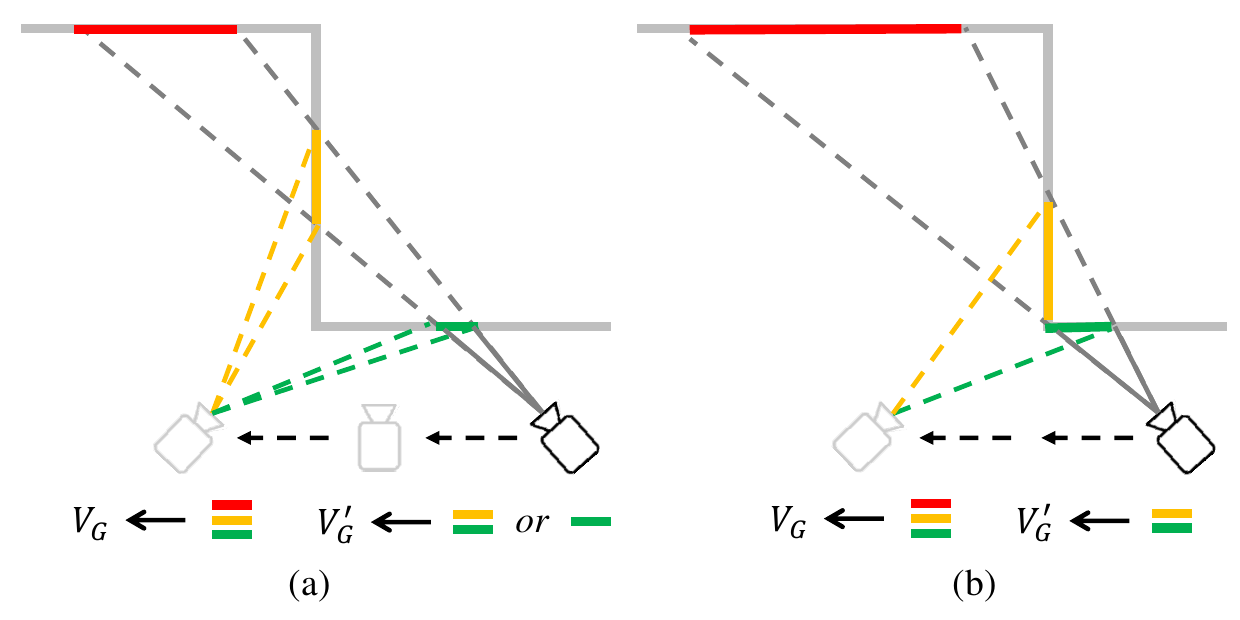} 
\end{center}
\vspace{-0.5cm}
\caption{Illustration of 2 possible situations when encountering occlusion. Here red region and yellow region are denoted as occluded points as they are not visible in the camera. In both situation, the red region will be filtered by clustering the depth information.}
\vspace{-0.4cm}
\label{vis3.2}
\end{figure}

When lifting 2D priors into 3D space, given a planar instance map $P_i$ with corresponding extrinsics $[R_i$, $t_i]$, and the intrinsic matrix $K$, we begin by projecting all nodes $V$ back into 2D camera coordinates. For each plane instance indicated in $P_i$, there is a group of points $V_G$ projected onto this region. We perform K-means clustering on projected depths with $K=2$ to coarsely filter out occluded points that may not appear in the image. An illustration figure of this process is shown in Fig.~\ref{vis3.2}. The occluded points will be projected onto the plane region together with the foreground points. We only consider the closest as plane-related points in each camera pose, so filtering out points with larger depth is necessary. Points with similar depths in one camera can be further distinguished through other views. The filtered point set is denoted as $V'_G$, and the edge $E(V_x, V_y \in V'_G)$ will be established among these points, as they are considered to be in the same plane from the plane instance $P_i$. For every two nodes $V_x, V_y \in V'_G$, the edge $E(V_x, V_y)$ will be created with the weight of 1 if it doesn't exist before. Otherwise, the weight will be incremented by 1. Using Leiden Algorithm to divide different communities, we identify the Gaussians distributed across each plane in the scene.

The ablation studies for hyperparameters $\alpha,\sigma$ are displayed in Tab.~\ref{tab:alpha} and Tab.~\ref{tab:sigma}. Here, we choose RaDe-GS as the baseline method, and run full settings of GSPlane. When implementing our experiments, we choose $\alpha=0.98$ and $\sigma=200$ as our settings.

\begin{table}[h]
\centering
\renewcommand{\arraystretch}{1}
\resizebox{0.6\columnwidth}{!}{%
\begin{tabular}{c|cccccc}
\hline
 $\alpha$ & Acc$\downarrow$   & Comp$\downarrow$  & Prec$\uparrow$  & Recall$\uparrow$ & F-score$\uparrow$ & \textit{num\_plane}\\ \hline

0.95     & \textbf{0.0821} & 0.0861 & 0.5168 & 0.672  & 0.5842 & 35.57   \\
\cellcolor{gray!20}0.98     & \cellcolor{gray!20}0.0824 & \cellcolor{gray!20}0.0855 & \cellcolor{gray!20}0.5197 & \cellcolor{gray!20}\textbf{0.6738}  & \cellcolor{gray!20}\textbf{0.5868} & \cellcolor{gray!20}34.43   \\ 
0.99     & 0.0831 & \textbf{0.0829} & \textbf{0.5214} & 0.6654  & 0.5846 & 31.29   \\ \hline
\end{tabular}%
}
% \vspace{-0.3cm}
\caption{Ablation on the cosine similarity threshold $\alpha$.}
% \vspace{-0.65cm}
\label{tab:alpha}
\end{table}

\begin{table}[h]
\centering
\renewcommand{\arraystretch}{1}
\resizebox{0.6\columnwidth}{!}{%
\begin{tabular}{c|cccccc}
\hline
 $\sigma$ & Acc$\downarrow$   & Comp$\downarrow$  & Prec$\uparrow$  & Recall$\uparrow$ & F-score$\uparrow$ & \textit{num\_plane}\\ \hline
100    & \textbf{0.0824} & \textbf{0.0855} & \textbf{0.5197} & \textbf{0.6738}  & \textbf{0.5868} & 34.43   \\
\cellcolor{gray!20}200     & \cellcolor{gray!20}\textbf{0.0824} & \cellcolor{gray!20}\textbf{0.0855} & \cellcolor{gray!20}\textbf{0.5197} & \cellcolor{gray!20}\textbf{0.6738}  & \cellcolor{gray!20}\textbf{0.5868} & \cellcolor{gray!20}34.43   \\
500    & 0.0827 & 0.0874 & 0.5175 & 0.6699  & 0.5839 & 32.14   \\\hline

\end{tabular}%
}
% \vspace{-0.3cm}
\caption{Ablation on the minimum pixel number $\sigma$ of K-means clustering.}
% \vspace{-0.65cm}
\label{tab:sigma}
\end{table}

\section{Algorithmic Illustration on Mesh Layout Refinement}\label{Appendix:mesh}

\begin{algorithm}
\caption{Mesh Layout Refinement}
\begin{algorithmic}[1]
\REQUIRE Extracted mesh $O$, Initial sparse point cloud $Pcd$, voxel size $\delta$, precomputed planar relationships $V_P^i \in Pcd$

\FOR{each plane $A \in V_P$}
    \FOR{each vertex $v_x \in O$}
        \IF{$\exists v_y \in V_P^A, |v_y - v_x| < 1.5\delta$ \textbf{and} $\forall \bar{v}_y \notin V_P^A, |\bar{v}_y - v_x| > 0.5\delta$}
            \STATE Assign $v_x$ to plane $A$ in $O$: $v_x \rightarrow \hat{V}_P^A \in O$
        \ENDIF
    \ENDFOR
\ENDFOR

\FOR{each $\hat{V}_P^A$}
    \STATE Remove planar faces: $\{f \in O | f = (v_1, v_2, v_3), v_1, v_2, v_3 \in \hat{V}_P^A\}$
    \STATE Categorize vertices: $\text{Boundary}~ \hat{V}_B^A, \text{Interior}~ \hat{V}_I^A$
    \STATE Project $\hat{V}_B^A, \hat{V}_I^A$ onto plane $A$: $\hat{V}_B^A \rightarrow \hat{V}_B^a$, $\hat{V}_I^A \rightarrow \hat{V}_I^a$
    \STATE Compute bounding rectangle $R_A$ covering $(\hat{V}_B^a, \hat{V}_I^a)$ and generate grid points $G_A$ within $R_A$
    \STATE Exclude $G_A$ points outside the projected region $(\hat{V}_B^a, \hat{V}_I^a)$
    \STATE Perform Delaunay triangulation: $T_A = \text{Delaunay}(V_B^A \cup G_A)$
    \STATE Map $T_A$ and $G_A$ back to 3D space
    \STATE Integrate $T_A,G_A$ into $O$
\ENDFOR

\RETURN Refined mesh $O'$
\end{algorithmic}
\label{alg:mesh_refinement}
\end{algorithm}

\clearpage

\section{Implementation of Dynamic Gaussain Re-classifier}\label{Appendix:B}

This section we provide some implementation details of our Dynamic Gaussian Re-classifier (DGR). The DGR is designed to identify and reclassify Gaussians that are mistakenly regarded as planar Gaussians. According to the general design of Gaussian training process, the distribution of Gaussians will be densified from Iteration 500 to 15,000 in each 100 iteration, and the whole training process will end at Iteration 30,000. Our DGR phase will be operating for the latter 50 iterations between every densification step, and for 100 iterations at Iteration 20,000. 

During the DGR phase, gradients of both planar Gaussians and non-planar Gaussians before finally proceeding to back-propagation will be stored and averaged for evaluation. The top 5\% of the planar gradients are selected and compared with the average magnitude of top 20\% non-planar gradients. Those with higher gradient magnitudes, the coordinates of their corresponding planar Gaussians will be re-formulated back to $xyz$ format. The DGR design can correct those mistaken planar Gaussians, and it will not influence the training for non-planar Gaussians. Thus, even if the true-positive planar Gaussians are processed, they will still be supervised with the baseline design.

\section{Additional Qualitative Results}\label{Appendix:C}

\begin{figure}[h]
\begin{center}
\includegraphics[width=0.8\linewidth]{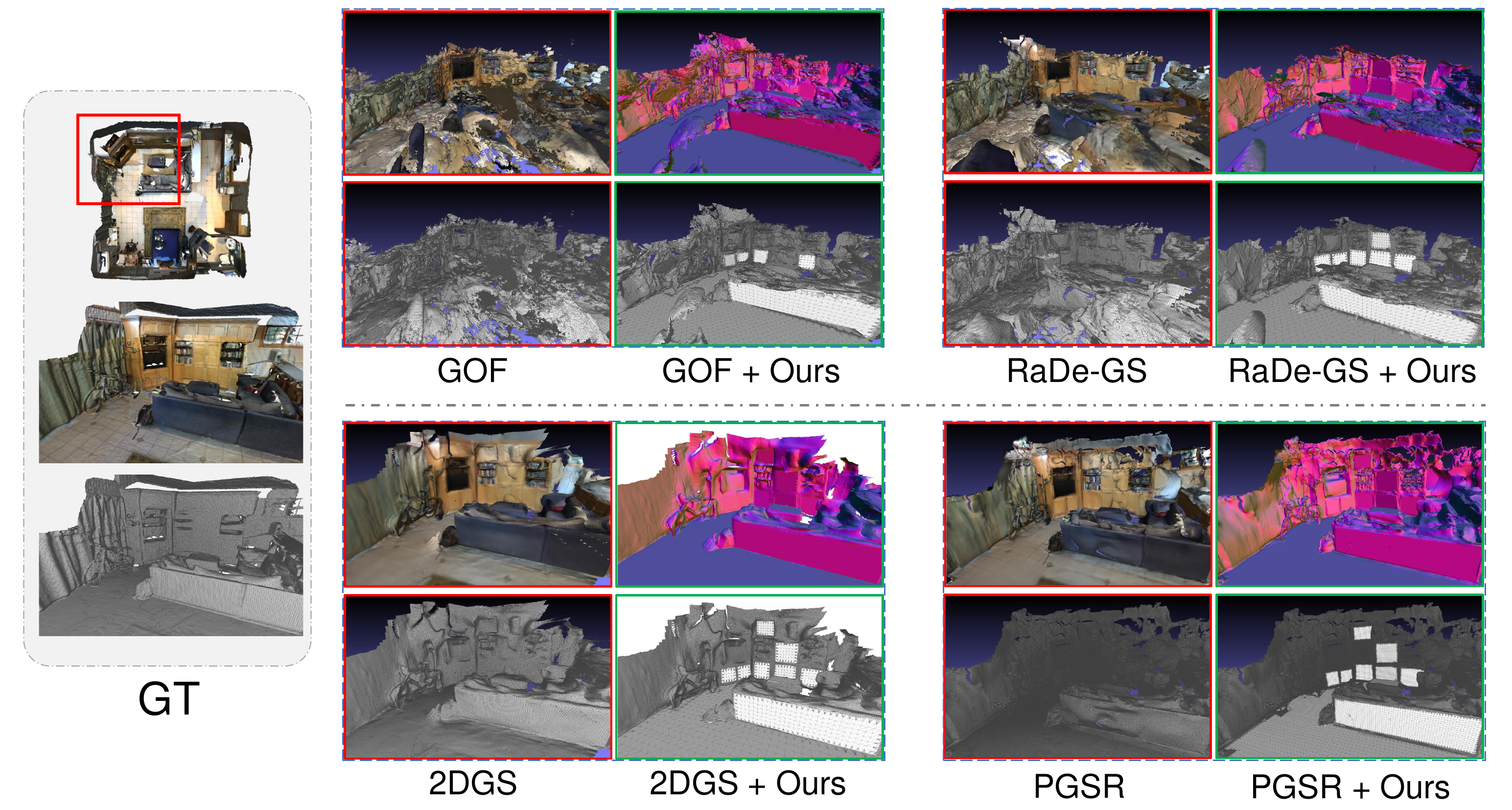} 
\end{center}
\vspace{-0.3cm}
\caption{Visualization of reconstructed mesh performance. }
\vspace{-0.5cm}
\label{appendix:vis}
\end{figure}

\begin{figure}[h]
\begin{center}
\includegraphics[width=0.8\linewidth]{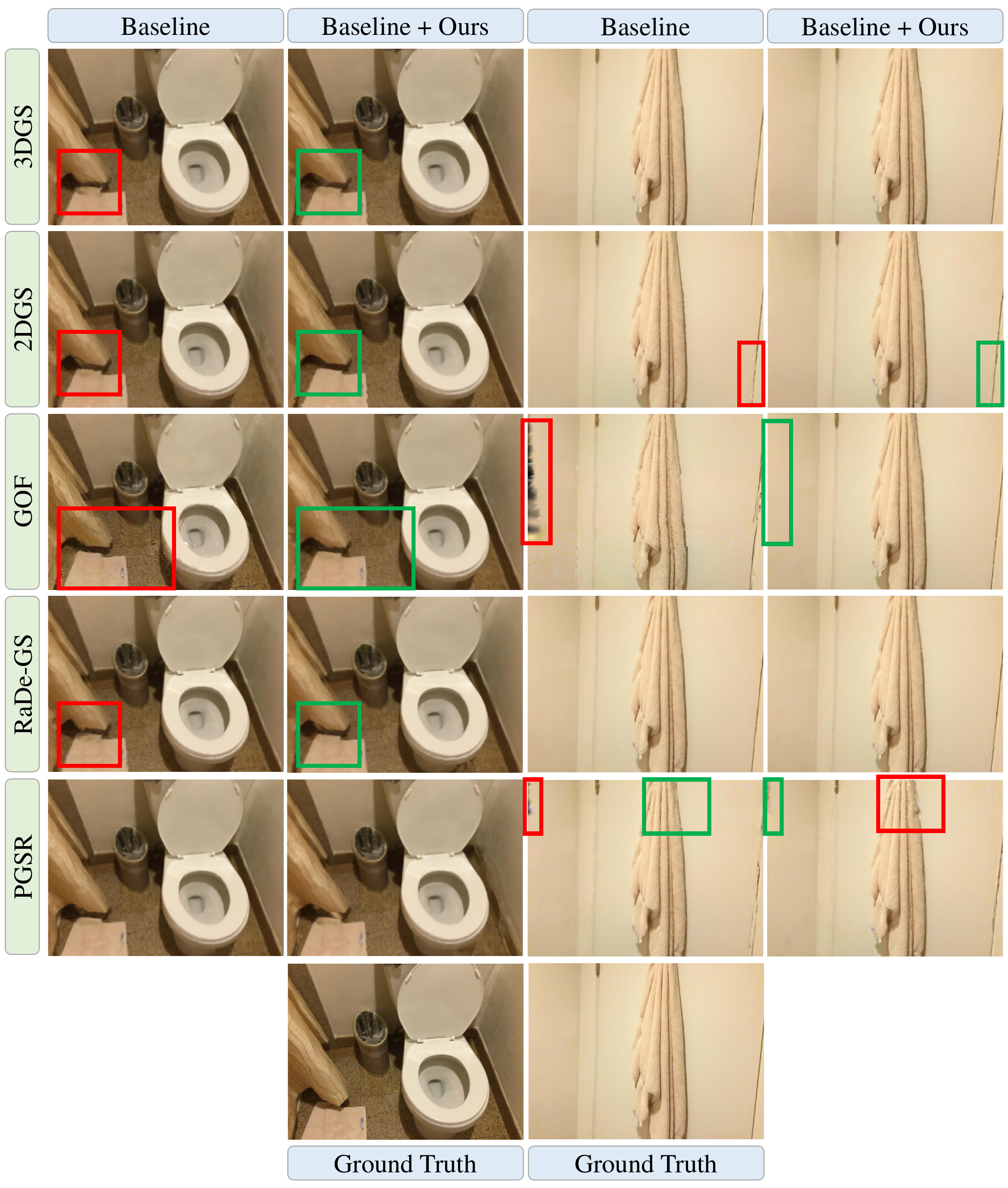} 
\end{center}
\vspace{-0.3cm}
\caption{Visualization of NVS results. }
\vspace{-0.5cm}
\label{nvs}
\end{figure}

In this section we first provide additional qualitative results on the overall reconstruction of the mesh in Fig.~\ref{appendix:vis}. We also provide examples in both rendering effects of GSPlane and baseline methods in Novel View Synthesis. The visualizations are shown in Fig.~\ref{nvs}. According to the quantitative results in Tab.~\ref{tab:1}, GSPlane also provides comparable results with small improvements, up to 0.018 and 1.02 for GOF in the SSIM and PSNR, respectively.

\section{Visualization of Planar Prior Extraction and Performance}\label{Appendix:D}

\begin{figure}[htbp]
    \centering
    \includegraphics[width=0.95\textwidth]{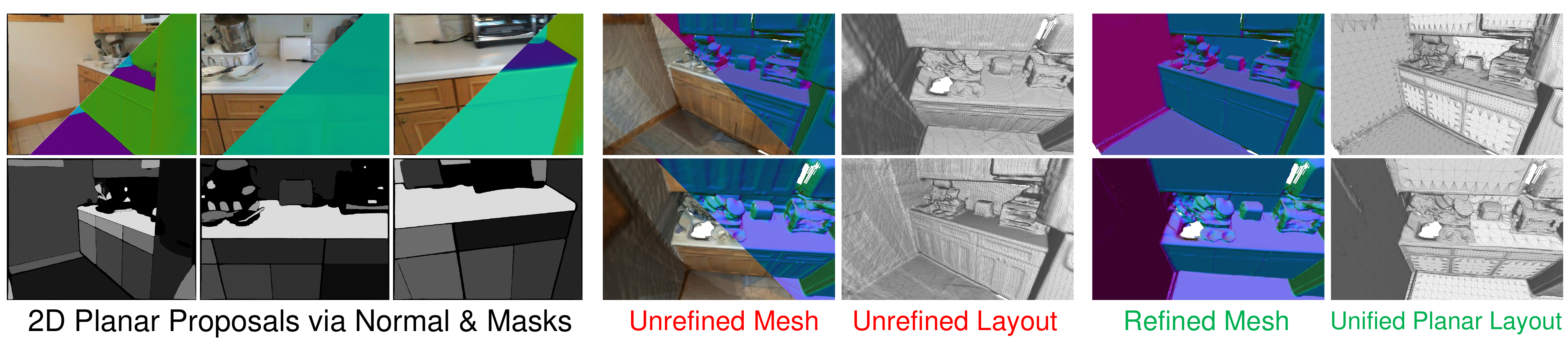}
    \caption{Visualization of an example from kitchen corner. The left shows the normal map and aggregated planar mask proposals of 2D views. The middle and right of the figure are the target mesh before \& after the layout refinement via structured representation of planes.}
    \label{fig:zoomin}
    \vspace{-0.3cm}
\end{figure}

In this section, we provide visualizations starting from 2D planar prior to the final refinement results in Fig.\ref{fig:zoomin}. Before training, we first establish planar priors by aggregating both subparts proposals from SAM~\cite{kirillov2023segment} and normal maps from Metric3Dv2~\cite{hu2024metric3d}. After structured representation for 3D planes are established, given a unrefined mesh with densely distributed vertices, GSPlane can create refined planar regions that exhibit consistent normals and topology, along with unified edges and a reduced number of vertices and faces, resulting in a more efficient and structured representation.

\clearpage
\section{Limitation}
Though GSPlane is able to provide concise and accurate geometry with satisfied topology and unified normal in planar region, there are still some issues before acquiring a desired and satisfied scene mesh. Currently, our focus is on planar regions, and the structured representation of non-planar regions remains an open challenge, which we leave as future work. A possible direction for addressing this issue could involve developing alternative representations tailored to complex surfaces. Additionally, the accuracy of planar priors are constrained by foundation models of masks and normals.

\section{LARGE LANGUAGE MODEL USAGE}

Large Language Models (LLMs) are used for polishing writing in this manuscript. The prompt is used as follows:

\textit{Assume you are a native English speaker, a senior researcher in the area of computer vision and graphics. Please help me polish the following content: \_\_\_}

\end{document}